\pdfoutput=1

\documentclass[10pt,twocolumn,letterpaper]{article}

\usepackage[pagenumbers]{cvpr} 

\usepackage[dvipsnames]{xcolor}
\definecolor{cvprblue}{rgb}{0.21,0.49,0.74}
\usepackage[pagebackref,breaklinks,colorlinks,citecolor=cvprblue]{hyperref}

\def\paperID{*****} 
\def\confName{CVPR}
\def\confYear{2024}

\usepackage[utf8]{inputenc} 
\usepackage[T1]{fontenc}    
\usepackage{hyperref}       
\usepackage{url}            
\usepackage{booktabs}       
\usepackage{amsfonts}       
\usepackage{nicefrac}       
\usepackage{microtype}      
\usepackage{xcolor}         

\usepackage{times}
\usepackage{epsfig}
\usepackage{graphicx}
\usepackage{amsmath}
\usepackage{amssymb}
\usepackage{mathabx}

\usepackage{bm}
\usepackage{xcolor}
\usepackage{verbatim}
\usepackage{multirow}
\usepackage{bbm}

\usepackage{caption}
\usepackage{soul}

\newcommand{\headline}[1]{\noindent \textbf{#1}}

\newcommand{\A}{\mathbf{A}}

\newcommand{\F}{\mathbf{F}}

\newcommand{\dF}{\check{\mathbf{F}}}

\newcommand{\Prob}{\mathbf{P}}

\newcommand{\mL}{\mathcal{L}}
\newcommand{\mR}{\mathbb{R}}

\newcommand{\y}{\boldsymbol{y}}

\renewcommand{\c}{\boldsymbol{c}}
\newcommand{\LM}{\mathbf{\Lambda}}

\newcommand{\attn}{\text{(attn)}}
\newcommand{\refine}{\text{(refine)}}

\newcommand{\mT}{\mathcal{T}}

\renewcommand{\Re}{\mathbb{R}}
\newcommand{\0}{\boldsymbol{0}}
\newcommand{\conv}{\operatorname{convolution}}
\newcommand{\tran}{\operatorname{transformer}}
\newcommand{\conc}{\operatorname{concat}}
\newcommand{\catt}{\operatorname{cross-attention}}
\newcommand{\amax}{\operatorname{argmax}}
\newcommand{\amin}{\operatorname{argmin}}
\newcommand{\dsample}{\operatorname{downsample}}
\newcommand{\usample}{\operatorname{upsample}}
\newcommand{\gru}{\operatorname{GRU}}
\newcommand{\fc}{\operatorname{FC}}
\newcommand{\bpi}{\boldsymbol{\pi}}

\title{BIT: Bi-Level Temporal Modeling for Efficient Supervised Action Segmentation}

\author{Zijia Lu\\
Northeastern Univeristy\\
{\tt\small lu.zij@northeastern.edu}
\and
Ehsan Elhamifar\\
Northeastern University\\
{\tt\small e.elhamifar@northeastern.edu}
}

\begin{document}

\maketitle

\begin{abstract}
We address the task of supervised action segmentation which aims to partition a video into non-overlapping segments, each representing a different action. Recent works apply transformers to perform temporal modeling at the frame-level, which suffer from high computational cost and cannot well capture action dependencies over long temporal horizons. To address these issues, we propose an efficient BI-level Temporal modeling (BIT) framework that learns explicit action tokens to represent action segments, in parallel performs temporal modeling on frame and action levels, while maintaining a low computational cost. Our model contains (i) a frame branch that uses convolution to learn frame-level relationships, (ii) an action branch that uses transformer to learn action-level dependencies with a small set of action tokens and (iii) cross-attentions to allow communication between the two branches. We apply and extend a set-prediction objective to allow each action token to represent one or multiple action segments, thus can avoid learning a large number of tokens over long videos with many segments. Thanks to the design of our action branch, we can also seamlessly leverage textual transcripts of videos (when available) to help action segmentation by using them to initialize the action tokens. We evaluate our model on four video datasets (two egocentric and two third-person) for action segmentation with and without transcripts, showing that BIT significantly improves the state-of-the-art accuracy with much lower computational cost (30 times faster) compared to existing transformer-based methods.

\end{abstract}

\section{Introduction}  

\begin{figure*}
   \centering
   \includegraphics[width=0.99\textwidth]{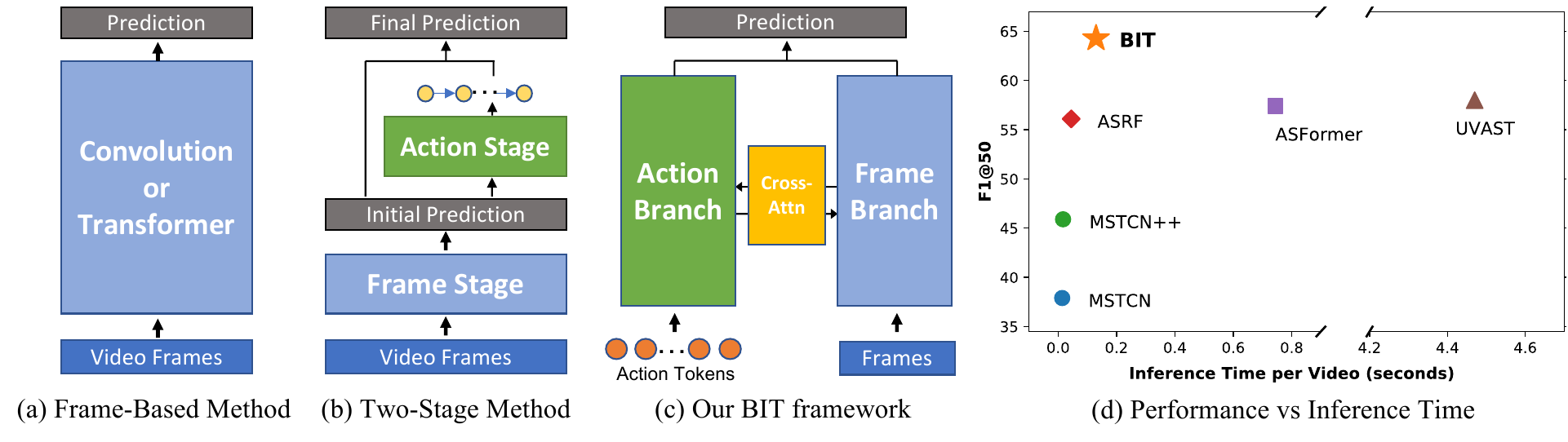}
   \caption{\small (a-c) \textbf{Architectures of prior and our methods}. (d) \textbf{Performance vs Inference Time} on Breakfast dataset: BIT outperforms prior methods while being \textit{30 times faster} than the previous best method (UVAST)}.
   \label{fig:intro}
   \vspace{-3mm}
\end{figure*}

Video action understanding, whose goal is to detect, recognize and segment human actions in a video, has broad applications in health, robotics, assistive technologies and other fields. 
In contrast to the action recognition task, which classifies the actions of short video clips, action segmentation aims to partition long and untrimmed videos into non-overlapping action segments\footnote{Action segment is a continuous clip of frames that represents one specific action in the video. For simplicity, we also refer to it as segment.} and has drawn increasing attention \cite{Farha:CVPR19,Li:TPAMI20,Huang:CVPR20,Ishikawa:WACV21,Yi:BMVC21,Dipika:Arxiv21,Ahn:ICCV21,Souri:PAMI21,Zhang:Arxiv22,Behrmann:ECCV22}.
A key challenge in this task is to understand the long temporal dependencies among actions, 
since many actions can only be accurately classified in the context of other actions (e.g., a person picking up a car wrench can proceed to loosen or tighten the lug nuts). 
Many existing works on action segmentation \cite{Farha:CVPR19,Li:TPAMI20,Ishikawa:WACV21,Yi:BMVC21,Dipika:Arxiv21} follow a frame-based model that estimates action relationships from frame features, see Figure \ref{fig:intro}(a). However, they have an inherent disadvantage in handling long videos where the model has to infer action relationships from tens of thousands of frames. 
Recent works \cite{Yi:BMVC21,Wang:Arxiv22} have employed transformers \cite{Vaswani:NIPS17} instead of temporal convolution \cite{Farha:CVPR19} to improve the segmentation performance, however, this comes with a great sacrifice in efficiency. 
Indeed, accurate modeling and inference of long-range action dependencies requires learning explicit representations of the action segments. However, this is a challenging task as it requires knowing the action segments of a video in advance, which is not available for test videos. Hence, \cite{Huang:CVPR20,Ahn:ICCV21,Behrmann:ECCV22} use a two-stage model that first computes initial framewise predictions then estimates action segments from the predictions, see Figure \ref{fig:intro}(b). 
Yet, they ignore the \textit{importance of bidirectional communication between the frame and action stages}. Their performance is limited by the quality of initial predictions, which is still obtained using the frame-based method.

\headline{Paper Contribution:}
%
To efficiently and accurately capture long-range temporal dependencies, we propose a \textbf{BI-level Temporal modeling (BIT)} framework that  
learns action tokens to represent action segments, estimates temporal relationships on frame and action levels \textit{in parallel} to allow easy cross-level communication, while maintaining a low computational cost. 
As shown in Figure \ref{fig:intro}(c),
BIT has a frame branch ({\color{blue} blue}) that operates on frame features to encode low-level details, an action branch ({\color{green} green}) that operates on a fixed-size set of action tokens for all videos to capture high-level action relationships, and cross-attentions ({\color{orange} yellow}) that enable communication between the two branches.
To infer a video, 
BIT assigns a subset of the tokens to encode the action segments in it while assigning the excessive ones to a special null class. 
We extend set-prediction loss \cite{Carion:CoRR20} to estimate the optimal matching between action tokens and ground-truth segments to supervise our model.
BIT has the following advantages over the state-of-the-art:

%

- Compared to frame-based models, BIT directly reasons about the dependencies among actions from their corresponding action tokens, rather than computing dependencies from the long sequence of frames. Therefore, it achieves \emph{more accurate predictions with less computations}. 

- Compared to two-stage methods, BIT learns the action tokens \emph{in parallel} to learning the frame features, while allowing them to leverage the \emph{complimentary information} in each other via cross-attention. It also contains multiple blocks to enable iterative refinement of the predictions.

- BIT is computationally efficient.
It uses \emph{transformer in the action branch} without a large computation overhead, as the number of action tokens is much smaller than the number of frames. 
It uses \emph{temporal convolution in the frame branch} to further reduce computation while frame features have access to long temporal information from the action tokens. 
BIT estimates all action tokens simultaneously, while prior work \cite{Behrmann:ECCV22} can only generate them auto-regressively. 

- Learning action tokens as a set instead of a sequence allows BIT to \emph{handle long videos with many action segments}, as there is no constraint on the mapping between action tokens and action segments. Thus, each token can represent one or multiple segments of the same action class.  
This effectively reduces the number of required tokens from being linear in the number of action segments to linear in the number of unique actions in a video. 

- BIT \emph{can leverage external knowledge} using the action tokens. 
Some videos (such as instructional videos) have textual narrations, which can be parsed into video transcripts describing the sequence of actions in videos \cite{Richard-Viterbi:CVPR18,Ding:CVPR18,Chang:CVPR19,Li:ICCV19,Lu:ICCV21,Souri:PAMI21,Richard:CVPR18,Li:CVPR20,Li:CVPR21-weaksup,Fayyaz:CVPR20, Shen:CVPR22,Li:CVPR21-timesup,Rahaman:ECCV22}. BIT can initialize the action tokens based on transcripts (when available), which improves the accuracy using even less training data. 

We extensively test BIT for action segmentation with and without transcripts on four video datasets, which range from third-person to egocentric videos, from small datasets with a few training videos to large datasets of long complex videos. 
As shown in Figure \ref{fig:intro}(d), \emph{BIT outperforms all prior methods while being 30 times faster than the previous state-of-the-art} \cite{Behrmann:ECCV22}.

\section{Related Works}
\vspace{-2mm}
\subsection{Action Segmentation} 
\vspace{-2mm}
Action segmentation has been studied in unsupervised \cite{Shen:CVPR21, Alayrac:CVPR16,Elhamifar:ICCV19,Zhukov:CVPR19,Kukleva:CVPR19,Fried:ACL20,Elhamifar:ECCV20}, weakly-supervised \cite{Richard-Viterbi:CVPR18,Ding:CVPR18,Chang:CVPR19,Li:ICCV19,Lu:ICCV21,Souri:PAMI21,Richard:CVPR18,Li:CVPR20,Li:CVPR21-weaksup,Fayyaz:CVPR20, Shen:CVPR22, Lu:CVPR22,Lu:ICCV21, Li:CVPR21-timesup,Rahaman:ECCV22} and full-supervised \cite{Farha:CVPR19,Li:TPAMI20,Yi:BMVC21,Dipika:Arxiv21,Behrmann:ECCV22,Ahn:ICCV21, Souri:PAMI21, Rohrbach:CVPR12, Singh:CVPR16, Kuehne:WACV16, Lea:CVPR17, Sigurdsson:CVPR17, Yeung:CVPR18,Zhang:Arxiv22,Li:CVPR22,Liu:Arxiv23} settings. BIT focuses on the fully-supervised setting. 
For frame-based methods, 
MSTCN \cite{Farha:CVPR19} and later works \cite{Li:TPAMI20,Dipika:Arxiv21} build models with temporal convolutions, which is computation efficient, yet their temporal receptive fields are limited by the number of layers. 
ASFormer \cite{Yi:BMVC21} replaces the convolution with transformer to improve performance. 
However, its computation is much higher than MSTCN (see Figure \ref{fig:intro}(d)) despite restricting one frame to attend to only a window of frames around it. 
\cite{Liu:Arxiv23} extends the multi-block refinement in ASFormer to a diffusion process, yet also leads to greater training and inference complexity. Since BIT focuses on improving temporal modeling while using a similar multi-block refinement, \cite{Liu:Arxiv23} is orthogonal to and can be incorporated in BIT.
On the other hand, Two-Stage methods \cite{Huang:CVPR20, Ahn:ICCV21, Behrmann:ECCV22} learn action segments based on initial framewise predictions and refine these predictions. However, the initial predictions are still obtained with a frame-based method and limits the performance.
Specifically, the best prior method, UVAST \cite{Behrmann:ECCV22}, uses transformer to compute framewise features, followed by a transformer decoder to predict the action segments. Yet it requires another alignment module, e.g., Viterbi decoding \cite{Forney:IEEE73,Richard-Viterbi:CVPR18}, to align the segments with the frames to obtain their locations, whose complexity is quadratic to the number of frames.
In contrast, BIT simultaneously learns temporal modeling on the video frames and action tokens while allowing communication between them using cross-attentions. 
Our cross-attentions obtain an accurate alignment between actions and frames, thus remove the need for additional alignment algorithms. 
\cite{Souri:PAMI21} explores a two-branch network in weakly-supervised setting, yet its goal is to apply cross-supervision between branches and has the same limitations as the two-stage methods.
While \cite{Xu:NIPS22} proposes a model-agnostic loss to enforce action orderings, our BIT method is a new model for more accurate and efficient temporal modelling and \cite{Xu:NIPS22} can be applied to BIT.
Lastly, \cite{Zhang:Arxiv22,Li:CVPR22} employ a pretrained vision-language model \cite{Wang:CoRR21} to enable prompt learning thus use more training data than our and other works.

\vspace{-2mm}
\subsection{Set Prediction}
\vspace{-2mm}
Set prediction methods \cite{Carion:CoRR20,Zhu:Arxiv20,Meng:ICCV21,Meinhardt:CVPR22, Dai:ICCV21} are recently introduced for the object detection task. 
DETR \cite{Carion:CoRR20} uses object tokens to encode the ground-truth objects in an image and applies a set prediction loss to find and learn the optimal one-to-one matching between them.
We firstly extend the set prediction method for the action segmentation task. 
One important motivation of using set is to efficiently handle long videos with many repeated segments. 
In fact, videos often contain repeated actions which have similar semantics and can be represented by a shared token.
For example, the longest video in the EPIC-Kitchen dataset \cite{Damen:IJCV22} contains 1140 action segments while 78\% of them are repeated.
\cite{Ahn:ICCV21,Behrmann:ECCV22} model segments by a sequence of action features, where the ordering indicates the matching between the features and the ground-truth segments. Thus, they only allow one-to-one matching as the ordering becomes ambiguous if one feature can correspond to several segments at different locations. 
In contrast, we construct the action tokens as a set, meaning there is no constraint on the matching from tokens to segments. Hence, we can match repeated segments to the same token and effectively reduce the number of required tokens, hence simplify the computation. 
While DETR \cite{Carion:CoRR20} addresses one-to-one matching, we propose a new algorithm to find one-to-many matching while ensuring all the matched segments of a token belong to the same action class.

\section{Proposed Method}

\begin{figure*}
   \includegraphics[width=1.0\textwidth]{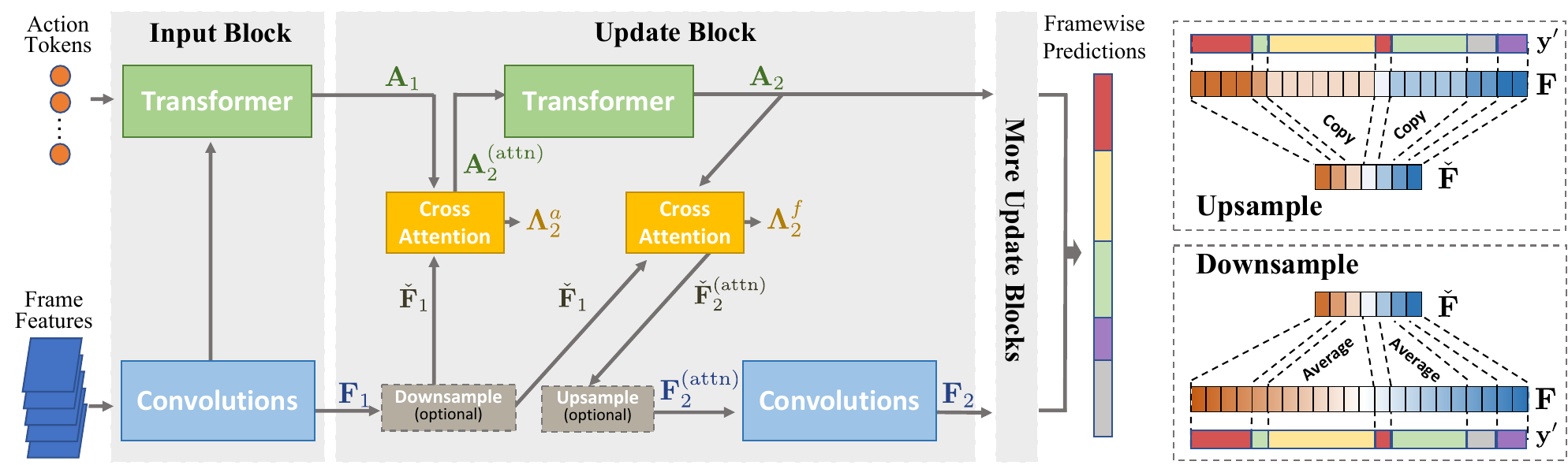}
   \caption{\small Our BIT network that learns explicit action tokens to represent action segments, performs temporal modeling on action and frame levels in parallel while maintaining low computation cost.}
   \label{fig:model}
\end{figure*}

We propose BIT framework for efficient supervised action segmentation. Given a test video with $T$ frames and pre-extracted\footnote{While similar to prior works we use pre-extracted features to save computation, BIT can be made end-to-end.} framewise features $\mathbf{X} \in \mathbb{R}^{T \times D}$, where $D$ denotes the feature dimension, our goal is to predict the action labels of all frames $\y \in [1, \ldots, A]^T $, where $A$ is the number of action classes. For training, we have videos and their ground-truth action segments.

\subsection{BIT Network Architecture}

\label{sec:model}
As shown in Figure \ref{fig:model}, our proposed architecture consists of an \emph{input block} and several \emph{update blocks}, where each block has a \emph{frame branch} ({\color{blue} bottom}) built using convolution to encode low-level details and an \emph{action branch} ({\color{green} top}) built using transformer to capture action dependencies. The input block performs initial feature learning for the two branches while update blocks enable communication between the two branches with cross-attention to refine the features.
Let $\A_0$ and $\F_0$ denote the inputs to the action and frame branches, respectively. 
We use a fixed small set of $M$ action tokens as $\A_0$, with a learned positional encoding $\rho^a$ to distinguish them, $\A_0, \rho^a \in \mR^{M \times D}$. 
We initialize the action tokens as $\A_0 = \0$ and our model updates the tokens to associate some of them to the ground-truth action segments while allowing others not to be assigned to any segment.
Depending on our loss function (see Section \ref{sec:loss}), one token can represent one segment (one-to-one) or multiple segments (one-to-many) of the same action class, while $M$ is set to be larger than the typical number of action segments or unique actions in a video, respectively. 
Action tokens learn the action classes and locations of their associated segments, while we do not allow two tokens to represent the same segment. 
For the frame branch, we use the framewise features $\F_0 = \mathbf{X}$ as the input with an absolute sinusoidal positional encoding $\rho^f$. 

\headline{Input Block:} We use the input block for initial feature learning of the action tokens and frame features.
We use multiple convolution layers to capture the temporal information among frames,
\begin{align}
(\F^{\refine}_1, &\Prob_1^f) = \conv(\F_0), \\
 \F_1 &= \conc(\F^{\refine}_1, \Prob_1^f),
\end{align}
where $\F^{\refine}_1$ is the updated frame features, $\Prob^f_1 \in \Re^{T \times A}$ is initial estimation of action probabilities for the frames and $\F_1$ is the concatenation of the two and also the output of the frame branch. 
We apply supervision on $\Prob^f_1$ to help the learning of $\F^{\refine}_1$, see Section \ref{sec:loss}. 

For the action branch, we use transformer with multi-head cross-attention and self-attention to initialize action tokens using frame features $\F_1$ and to learn the dependencies among tokens,
\begin{align}
(\A^\refine_1, \Prob^a_1) &= \tran(\A_0 + \rho^a; \F_1 + \rho^f), \\ 
\A_1 &= \conc(\A^\refine_1, \Prob_1^a).
\end{align}
Similarly, $\A^\refine_1$ is the updated features of action tokens, $\Prob^a_1 \in \mathbb{R}^{M \times A+1}$ is the action probabilities of tokens, hence the action probabilities for their associated segments, and $\A_1$ is the output of the action branch. Similar to \cite{Carion:CoRR20}, in addition to the $A$ real action classes, we include a special null class that represents the label of tokens not assigned to any segment.

\headline{Update Block:} 
The purpose of our update block is to use cross-attention to allow frame features leveraging the high-level action dependencies learned in the action branch and allow action tokens accessing the low-level information in the frame branch. 

To start, we perform cross-attention with only one attention head to update action tokens based on frame features, i.e., using $\A_1$ as queries and $\F_1$ as keys and values,
\begin{align}
(\A^\attn_2, \mathbf{\Lambda}^a_2) = \catt(\A_1 + \rho^a; \F_1 + \rho^f),
\label{eq:f2a}
\end{align}
where $\A^\attn_2$ denotes the updated action tokens and $\LM^a_2 \in \mR^{M \times T}$ is the obtained one attention map. Each row of $\LM^a_2$ denotes the attention from a token to all the frames and will sum to one.
We use one attention map as it allows us to obtain an alignment between action tokens and frames (indicating the temporal locations of the associated segments of the tokens). 
Next, we use transformer with only multi-head self-attention to refine the tokens and predict their action classes, 
\begin{align}
(\A^\refine_2, \Prob^a_2) &= \tran(\A^\attn_2 + \rho^a), \\
 \A_2 &= \conc(\A^\refine_2, \Prob^a_2), 
\label{eq:a-update}
\end{align}
where $\A^\refine_2, \Prob^a_2, \A_2$ are, respectively, the refined action tokens, probabilities of their action classes and the output of the action branch.
%

Given the updated output of the action branch, $\A_2$, we similarly use single-head cross-attention to update the frame features and refine them with convolutions, 
\begin{align}
(\F^\attn_2, \mathbf{\Lambda}^f_2) &= \catt(\F_1 + \rho^f; \A_2 + \rho^a), \label{eq:a2f} \\
(\F^\refine_2, & \Prob^f_2)  = \conv(\F^\attn_2), \\
\F_2 & =  \conc(\F^\refine_2, \Prob^f_2), \label{eq:f-update}
\end{align}
where $\mathbf{\Lambda}^f_2 \in \Re^{M \times T}$ is the attention from frames to action tokens (each column sums to one). It also indicates the alignment between action tokens and frames based on the latest features. 
$\Prob^f_2$ is the updated estimation of the actions of frames. 
Finally, $\A_2$ and $\F_2$ are inputs to the next update block.

\headline{Temporal Downsampling: } 
When there is a large difference between the number of actions and frames, learning cross-attention becomes challenging (as also observed in \cite{Behrmann:ECCV22}, hence, resorting to the alternative yet costly strategy of Viterbi decoding performs better). 
We show that this issue can be effectively resolved by properly downsampling the frame features temporally before cross-attention, followed by upsampling them. More specifically, as shown in Figure \ref{fig:model} (left), we compute downsampled features $\check{\F}_1$ to replace $\F_1$ in \eqref{eq:f2a} and \eqref{eq:a2f}.  
From \eqref{eq:a2f} we obtain $\dF^\attn_{2}$, which is the output of cross-attention. We then upsample it to obtain $\F^\attn_2$ to use it in \eqref{eq:f-update}.

We do not downsample $\F_1$ by a fixed ratio, as it removes features of short segments.
Instead, we partition the video into segments based on action predictions of the frames, $\y' = \amax(\Prob^f_1)$, then downsample $\F_1$ by computing one feature for each segment via average-pooling, as shown in Figure \ref{fig:model} (right).
We found it is helpful to further refine the obtained features through GRU. 
Thus, we have $\check{\F}_1 = \gru(\dsample(\F_1))$. The positional encodings of $\dF_1$, $\check{\rho}^f$, are the positional encodings of the middle frames of the segments.
Next, to upsample $\dF^\attn_{2}$, we make copies of its features, see Figure \ref{fig:model} (right). Since the upsampled features lose low-level details of frames, we merge it with $\F_1$ using a fully-connected layer, i.e., $\F^\attn_2 = \fc(\F_{1}, \usample(\dF^\attn_{2}))$, which we use in \eqref{eq:f-update}. 
Notice that the attention maps obtained in this process are also downsampled, hence, we will similarly upsample them for later usage (see supplementary materials for more details).

\headline{Generating Predictions: }
BIT consists of $B$ blocks, with 1 input and $B-1$ update blocks. Using the last update block, we can generate action predictions for frames using i) the output of the frame branch by computing $\amax(\Prob^f_B) \in [1, \ldots, A]^T$, or ii) the output of the action branch. 
To do so, we compute the predicted action classes of action tokens, $\c = \amax(\Prob^a_B) \in [1, \ldots, A+1]^M$, and the frames assigned to tokens  using the last cross-attention, $m_t = \amax(\LM^f_B(m, t))$. This means frame $t$ belongs to the segment of token $m_t$ thus its predicted class is $\c(m_t)$.
Notice that, for tokens classified to the null class, we mask out their attentions in $\LM^f_B$ to avoid assigning frames to them. 
We use the average of the two predictions for final prediction.

\subsection{Proposed Loss Functions}
\label{sec:loss}

We supervise the frame branch with the ground-truth $\y$ to learn the action classes of frames. 
For the action branch, action tokens need to correctly learn the action classes and locations of the ground-truth segments. 
Since we construct the tokens as an unordered set, there is no predefined matching between tokens and segments. Thus, we search for an optimal matching between them and use it as supervision to compute our losses.  
Assuming there are $N$ ground-truth segments, segment $n$ can be described by its action class $a_n$ and its temporal interval $\mathcal{T}_n$. 
We define the optimal token-segment matching as $\bpi^* \in [1, \ldots, M ] ^N$, where $\bpi^*_n = m$ means that segment $n$ is assigned to token $m$. Next, we first discuss our loss functions then discuss the procedure for finding $\bpi^*_n$ at the end of this subsection. 

\headline{Frame Loss} enforces the action probabilities of the frames, $\Prob^f_b$, to conform with $\y$,
\begin{align}
    \mL_\text{f} = \sum_{b} \frac{1}{T}\sum_{t} - \log \Prob^f_b(t, \y_t),
\end{align}
where $\Prob^f_b(t, \y_t)$ is the probability that frame $t$ belongs to class $\y_t$, obtained from the $b$-th block.

\headline{Action Token Loss} enforces that action tokens learn the action classes of their associated segments,
\begin{align}
   \mL_\text{a} = \sum_{b} \frac{1}{M}\Big[ - \sum_{n} \log \Prob^a_b(\bpi^{*}_n, a_n) - \sum_{m \in \mathcal{N}} \log \Prob^a_b(m, A+1) \Big],
\end{align}
where the first term requires the tokens to have the same classes as their matched segments. 
The second term requires the tokens belong to the null class if they are not matched to any segment, where $\mathcal{N} = \{m | m \notin \bpi^*\}$ stores the indices of those tokens.

\headline{Cross-Attention Loss} enforces that tokens attend to the frames of their matched segments, while frames in each segment attend to their matched token,
\begin{align}
   \mL_\text{c} = \sum_{b > 1} \frac{1}{T} \sum_{n} \sum_{t \in \mT_n} - \left( \log \LM^a_b(\bpi^{*}_n, t) + \log \LM^f_b(\bpi^{*}_n, t) \right),
\end{align}
where the two terms are the attention weights between a frame $t$ in segment $n$ and the matched token of segment $n$, i.e., $\bpi^*_n$.
We apply no constraint on the attentions of the tokens of null class.
Notice $\mL_\text{c}$ is not applied to the input block, since there is no cross-attention in the input block. 

\headline{Temporal Smoothing Loss} addresses the over-segmentation issue where frawewise predictions oscillate between actions around action boundaries. We apply the smoothing loss on the action probabilities of frames and alignment (attention map) between frames and action tokens,
\begin{align}
    \mL_\text{ts} = w \sum_{b} h(\Prob^f_b) + h(\LM^a_b) + h(\LM^f_b),
\end{align}
where $h(\cdot)$ is a smoothing loss \cite{Farha:CVPR19} (see the supplementary material for details) and $w$ is the weight, putting a trade-off with other losses.
Finally, our overall loss is $\mL = \mL_{a} + \mL_{f} + \mL_{c} + \mL_{ts}$. 

\headline{Computing Optimal Token-Segment Matching.} To obtain the optimal matching $\bpi^*$, let $S(n, m)$ be the matching cost between segment $n$ and token $m$. 
The optimal matching minimizes the total matching cost, $\bpi^{*} = \amin_{\bpi} \sum_{n} S(n, \bpi_n)$, subject to one-to-one or one-to-many matching constraints. 
Specifically, when each token can represent one segment (one-to-one), we enforce that each $m$ does not appear in $\bpi^{*}$ more than once and solve it via the Hungarian algorithm \cite{Kuhn:Hungarian}. When each token can represent multiple segments (one-to-many), we require the segments must belong to the same action class and solve it via a proposed three-step algorithm, which we discuss in the supplementary materials due to lack of space. 

The matching cost $S(n, m)$ will take into account both the predicted action class of token $m$ and the frames it is associated with in the cross-attention and is defined as
\begin{align}
    S(n, m) =  - \, \Prob^a_B(m, a_n) 
    - \beta \, 
        \frac{\sum_{t} \LM^f_B(m, t) \cdot \operatorname{1}_{t \in \mT_n}}
    {\sum_{t} \min(\LM^f_B(m, t) + \operatorname{1}_{t \in \mT_n}, 1)},
    \label{eq:similarity}
\end{align}
where the first term, $\Prob^a_B(m, a_n)$, is the probability that the token belongs to the same action as the segment. 
The second term is a soft IoU score between the segment $m$ and the frames that the token is associated with.
$\operatorname{1}_{t \in \mT_n}$ is an indicator function that is 1 if $t$ resides in the segment and 0 otherwise. 
$\LM^f_B(m, t) \in [0, 1]$ is the $(m, t)$-th entry of the matrix and indicates if frame $t$ is aligned to token $m$. 
Lastly, $\beta$ controls the balance between these two terms.

 \subsection{Leveraging Video Transcripts}
\label{sec:action-seg-w-transcript}
BIT can also easily incorporate video transcripts (an ordered list of the segments in a video). Notice that the temporal locations of segments are still unknown. 
When the transcripts are available, we use them to initialize action tokens. 
Specifically, 1) we no longer learn a set of action tokens but learn embeddings of the action classes. The action tokens are constructed as the embeddings of the actions in the transcript, with an absolute sinusoidal positional encoding.  
2) creating the tokens based on the transcript means we know the ground-truth matchings between the tokens and segments, 
thus we can use it in our losses to replace the optimal matching $\bpi^*$. 
With these two simple changes, BIT can incorporate textual transcripts while maintaining the same inference speed. This is in contrast to prior works \cite{Farha:CVPR19,Li:TPAMI20,Yi:BMVC21,Dipika:Arxiv21,Ahn:ICCV21} that do not directly use transcripts and can only be extended via post-processing, which greatly increases the computational time.

\section{Experiments}
We evaluate our method for action segmentation on four challenging datasets and compare with prior methods \cite{Farha:CVPR19,Li:TPAMI20,Ishikawa:WACV21,Yi:BMVC21,Behrmann:ECCV22}, specifically with ASFormer\cite{Yi:BMVC21} and UVAST\cite{Behrmann:ECCV22}, which are the best frame-based and two-stage methods, respectively.

\subsection{Experimental Setup}
\headline{Datasets.} We evaluate on four datasets, representing different test scenarios: 
\textit{Breakfast} \cite{Serre:CVPR14} is a third-person cooking dataset and contains 1716 videos from 10 recipes and 48 actions with an average of 6.9 action segments per video.
\textit{GTEA} \cite{Fathi:CVPR21} is a small-scale dataset with 28 videos to test \textit{learning with limited data}. It has 11 actions and on average 33 segments per video.
\textit{EgoProceL} \cite{Bansal:ECCV2022} is an egocentric dataset featuring \textit{diverse tasks}, such as repairing cars, assembling toys and cooking. 
It has 1055 videos, 130 actions and on average 21 segments per video.
\textit{EPIC-Kitchen} \cite{Damen:IJCV22} is the most challenging dataset featuring \textit{long complex videos}. It has 633 videos, \textit{3796} actions and on average 195 segments per video, while the longest video contains 1436 segments. 

\headline{Metrics.}
Following prior works \cite{Farha:CVPR19,Li:TPAMI20,Yi:BMVC21,Dipika:Arxiv21,Behrmann:ECCV22}, we compute segmental Edit distance score (\textit{Edit}) and segmental F1 score (\textit{F1}) at three overlapping threshold 10\%, 25\%, 50\%, denoted by F1@\{10, 25, 50\}. 
\textit{Edit} measures if the action sequence in the framewise prediction matches with that of the ground-truth segments without considering their temporal locations and durations. 
\textit{F1} measures if an action segment is correctly detected based on its IoU with the ground-truth segments. 
We also measure the frame-wise accuracy (\textit{Acc}).

\headline{Implementation.}
We learn our model with 1 input block and 3 update blocks with the dual dilated layers from \cite{Li:TPAMI20} for convolution.
We learn one-to-one matching between action tokens and segments on Breakfast, GTEA and EgoProceL with 60, 60 and 200 tokens, respectively, and one-to-many matching on EPIC-Kitchen with 300 tokens, since learning one-to-one matching requires at least 1500 tokens, which would be less efficient. We apply the temporal downsampling in the later blocks of our network (see supplementary materials for details). 
On EgoProceL and EPIC-Kitchen, we reproduce the best prior works, ASFormer, UVAST, and other methods \cite{Farha:CVPR19,Li:TPAMI20,Ishikawa:WACV21} using their released codes, as they do not report on the two datasets in their papers.
The result of UVAST is not reported on EPIC-Kitchen as we found its sequence decoder has difficulty learning the large number of segments in the videos, thus cannot converge well.
We include more implementation details in the supplementary materials \footnote{We will release our code and model weights.}.

\subsection{Comparison with the State-of-the-Art}

\headline{Action Segmentation.}
Table \ref{table:aseg-no-trans} shows the results of different action segmentation methods. 
\textbf{We achieved new state-of-the-art results on all datasets on all metrics}, exceeding the best prior work on F1@50 by \textit{6.7}\%, \textit{1.6}\%, \textit{7.5}\% and \textit{6.1}\% on Breakfast (widely-used benchmark), GTEA (small-scale), EgoProceL (diverse tasks) and EPIC-Kitchen (long complex videos), respectively. 
More importantly, as shown in Figure \ref{fig:intro}(d), on Breakfast \textbf{the inference time of BIT is \textit{30} times faster than UVAST and \textit{6} times faster than ASFormer}.
On EPIC-Kitchen, while UVAST cannot converge well due to the large amount of action segments per video, our one-to-many matching allows using less action tokens than the number of action segments thus reduces the learning difficulty, the efficacy of which is validated in the improved F1.

\begin{table*}[]
   \fontsize{8.6pt}{11pt}\selectfont
   \centering
\addtolength{\tabcolsep}{-2.5pt}
   \begin{tabular}{l|ccc|c|c|ccc|c|c|ccc|c|c|ccc|c|c|}
   \hline
   \hline
   & \multicolumn{5}{|c|}{Breakfast} & \multicolumn{5}{|c|}{GTEA} & \multicolumn{5}{|c|}{EgoProceL} & \multicolumn{5}{|c|}{Epic-Kitchen} \\ \cline{2-21}
   & \multicolumn{3}{c|}{ F1@\{10,25,50\}} & { Edit} & { Acc}&
                     \multicolumn{3}{c|}{ F1@\{10,25,50\}} & { Edit} & { Acc} & 
                        \multicolumn{3}{c|}{ F1@\{10,25,50\}} & { Edit} & { Acc} & 
                        \multicolumn{3}{c|}{ F1@\{10,25,50\}} & { Edit} & { Acc} \\ \hline

   {MSTCN++\cite{Li:TPAMI20}} & 64.1 & 58.6 & 45.9 & 64.9 & 67.6
   & 88.8 & 85.7 & 76.0 & 83.5 & 80.1
   & 60.3 & 57.0 & 46.5 & 62.4 & 69.3
   & 15.2 & 13.6 & 9.5 & 11.6 & 18.2  \\
   {ASRF\cite{Ishikawa:WACV21}} & 74.3 & 68.9 & 56.1 & 72.4 & 67.6
   & 89.4 & 87.8 & 79.8 & 83.7 & 77.3
   &  - & - & - & - & - 
   &  - & - & - & - & - \\
   {ASFormer\cite{Yi:BMVC21}} & 76.0 & 70.6 & 57.4 & 75.0 & 73.5
   & 90.1 & 88.8 & 79.2 & 84.6 & 79.7
   & 63.3 & 60.9 & 51.0 & 64.9 & 71.1
   & 16.4 & 14.8 & 10.5 & 12.6 & 19.1 \\
   {UVAST\cite{Behrmann:ECCV22}} & 76.9 & 71.5 & 58.0 & 77.1 & 69.7
   & 92.7 & 91.3 & 81.0 & 92.1 & 80.2 
   & 60.6 & 56.1 & 48.3 & 71.9 & 69.6 
   &  - & - & - & - & - \\ \hline
   {\textbf{BIT}}  & \textbf{80.6} & \textbf{75.9} & \textbf{64.7} & \textbf{79.0} & \textbf{75.5}
   & \textbf{94.8} & \textbf{92.8} & \textbf{82.6} & \textbf{92.6} & \textbf{82.0}
   & \textbf{76.9} & \textbf{74.1} & \textbf{64.0} & \textbf{79.2} & \textbf{88.0} 
   & \textbf{35.9} & \textbf{30.6} & \textbf{20.1} & \textbf{37.2} & \textbf{31.3} \\

   \hline
   \end{tabular}
   \caption{Action Segmentation Performance.} 
   \label{table:aseg-no-trans}

   \vspace{1mm}

   \begin{tabular}{l|ccc|c|c|ccc|c|c|ccc|c|c|ccc|c|c|}
    \hline
    \hline
    & \multicolumn{5}{|c|}{Breakfast} & \multicolumn{5}{|c|}{GTEA} & \multicolumn{5}{|c|}{EgoProceL} & \multicolumn{5}{|c|}{Epic-Kitchen} \\ \cline{2-21}
    & \multicolumn{3}{c|}{ F1@\{10,25,50\}} & { Edit} & { Acc}&
                      \multicolumn{3}{c|}{ F1@\{10,25,50\}} & { Edit} & { Acc} & 
                         \multicolumn{3}{c|}{ F1@\{10,25,50\}} & { Edit} & { Acc} & 
                         \multicolumn{3}{c|}{ F1@\{10,25,50\}} & { Edit} & { Acc} \\ \hline
 
   MSTCN++\cite{Li:TPAMI20} & 84.0 & 75.7 & 59.4 & - & 72.2
      & 94.3 & 90.5 & 78.9 & - & 78.3  
      & 64.0 & 57.2 & 46.4 & - & 69.9
      & 55.0 & 42.5 & 22.9 & - & 28.1 \\
   ASFormer\cite{Yi:BMVC21} & 85.4 & 78.5 & 63.9 & - & 74.6
        & 95.8 & 94.1 & 82.9 & - & 82.0
        & 64.2 & 58.5 & 47.6 & - & 72.0
        & 58.2 & 45.5 & 24.6 & - & 32.1 \\
   UVAST\cite{Behrmann:ECCV22} & 87.6 & 81.9 & 69.2 & - & 77.0
        & \textbf{96.9} & 95.3 & 86.0 & - & 82.7
        & 68.5 & 63.6 & 50.6 & - & 75.8
        &  - & - & - & - & - \\ \hline
   \textbf{BIT} & \textbf{89.9} & \textbf{85.6} & \textbf{73.7} & \textbf{93.5} & \textbf{84.5}
               & 96.1 & \textbf{95.6} & \textbf{87.5} & \textbf{96.3} & \textbf{84.0}
               & \textbf{74.3} & \textbf{71.2} & \textbf{61.9} & \textbf{87.2} & \textbf{80.5}
               & \textbf{66.9} & \textbf{60.8} & \textbf{44.8} & \textbf{75.2} & \textbf{58.2} \\
 
    \hline
    \end{tabular}
    \caption{Action Segmentation Performance using Video Transcripts} 
    \label{table:aseg-wt-trans}

\end{table*}

\headline{Action Segmentation with Transcripts.} When video transcripts are available during training and testing, action segmentation can be considered as learning alignments between frames and the action segments given by the transcripts. 
Thus, we extend prior works by applying Viterbi decoding \cite{Forney:IEEE73} to obtain the optimal alignments between their framewise predictions and the transcripts, which has shown reliable performance in many prior works \cite{Richard-Viterbi:CVPR18,Li:CVPR21-weaksup,Lu:ICCV21}. 
Since it ensures their predictions obey the transcripts, the Edit score will always be 100 and cannot reflect the model performance. 

As Table \ref{table:aseg-wt-trans} shows, despite not using Viterbi decoding, BIT still achieves the best performance, improving F1@50 by \textit{4.5\%, 1.5\%, 7.8\%, 18.6\%} on Breakfast, GTEA, EgoProceL and EPIC-Kitchen, respectively. 
The high Edit scores also indicate the predictions of BIT closely align with the transcripts. 
It is because \emph{BIT can use transcripts to initialize action tokens and improve feature learning} while prior works can only apply Viterbi decoding as a post-processing step.  
Interestingly, ASFormer has lower F1@50 compared to the results without transcript in Table \ref{table:aseg-no-trans}. 
It is because, when a ground-truth segment is missing in its predictions, Viterbi decoding infers the location of the segment, which can incorrectly modify the locations of other correct segments. 
In contrast, including transcript can consistently improve BIT on all datasets.
We report the results of other baselines in the supplementary materials.

\subsection{Ablation Studies}
We test the effect of the number of action tokens, different matching types between action tokens and segments, each proposed loss function and incorporating video transcripts. 

\vspace{1mm}

\begin{figure}
   \centering
   \includegraphics[width=0.45\textwidth]{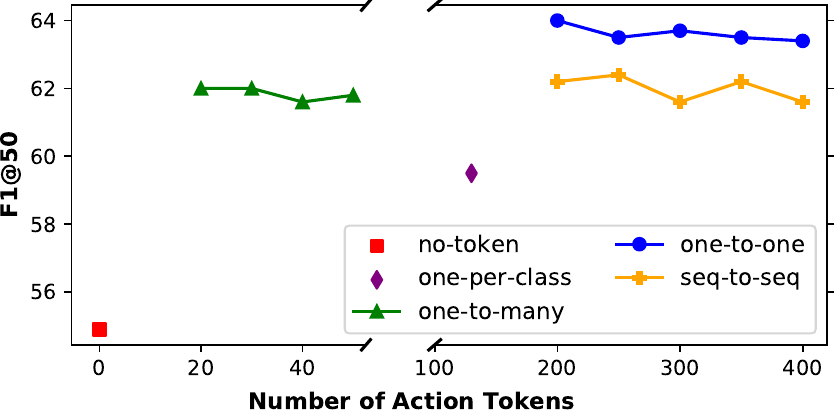}
   \captionof{figure}{\small Effect of Action Token.}
   \label{fig:action-token}
\end{figure}

\headline{Number of Action Tokens.}
In Figure \ref{fig:action-token}, we test learning different number of action tokens on EgoProceL for action segmentation. 
First, \textit{no-token} shows a baseline model that only has the frame branch. Thus, it becomes a frame-based method and cannot well capture the long temporal dependency, leading to a very low F1.  
On the other hand, using action tokens with \textit{One-to-Many} (OTM) or \textit{One-to-One} (OTO) matching between tokens and segments improves F1 by 8-10\%, showing action-level temporal modeling is key to good action segmentation. Moreover, the accuracies of both OTO and OTM are robust to the number of tokens, thus we can choose to use fewer tokens without performance sacrifice.

\headline{Matching between Action Tokens and Segments.} In Figure \ref{fig:action-token}, we also show the effect of learning using different matchings between action tokens and segments. 
First, OTO matching shows better F1 than OTM as it separately represents each segment thus better encodes their action classes and locations, yet the performance gap between the two is not large. 
We also compare OTM with a \emph{One-Per-Class} (OPC) matching, where we always assign a token to encode the segments from a specific action class. 
Notice the number of required tokens for OPC is the number of \textit{all} action classes and is more than that of OTM, which is linear to the number of action classes within one video. 
OPC also has a lower F1 as it always fuses the information of segments of the same action, while OTM allows the model to decide when to encode segments with one token. 
Lastly, we compare OTO with \textit{Seq-To-Seq} (STS), where we consider the tokens as a sequence, matching the first $N$ tokens to the $N$ segments (recall that $N$ is the number of ground-truth segments) and the rest to the null class. STS obtains lower F1 because it imposes sequential dependency between tokens, thus mistakes in one token (e.g., token $5$ should predict the $5$-th segment but incorrectly predicts the $6$-th segment) affects the predictions of all the subsequent tokens. In OTO, such error in one token will not affect the others.

\headline{Effect of Losses}
In Table \ref{table:losses}, we show the effect of our proposed losses on Split 1 of Breakfast. 
Removing $\mL_\text{f}$ (first row) leads to degradation of the frame features, which also affects the action tokens, hence causes large performance drop.  
%
When we remove supervision for action branch (second and third rows), it harms the model's ability to capture long temporal relations, leading to drop in all metrics.
Removing $\mL_\text{ts}$ (forth row) leads to over-segmentation, where the model predicts many short false positive segments, thus decrease F1 and Edit.

\begin{figure}
   \centering
   \includegraphics[width=0.45\textwidth]{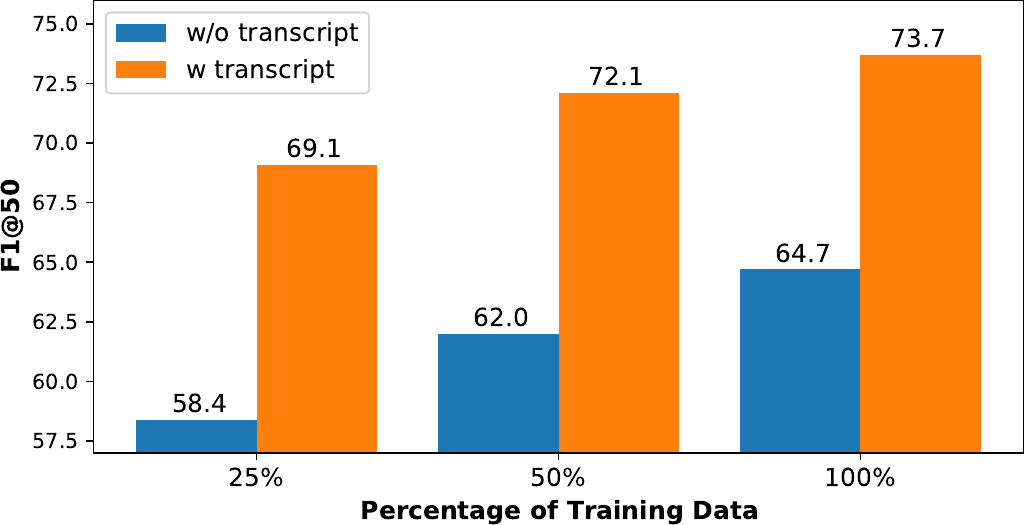}
   \captionof{figure}{\small Learning with Less Videos.}
   \label{table:data}
\end{figure}

\headline{Learning with less data.}
One important motivation for incorporating transcript is to leverage the information from textual modality to more effectively learn a model with less data and annotations. To valid this, we test learning with 25\%, 50\% and 100\% training data on Breakfast. As Figure \ref{table:data} shows, with just 25\% training data, our model with transcripts outperforms the one not using transcripts but with 100\% training data. \emph{This shows our model can be applied to scenarios with a small number of training videos or where annotation is sparse}. 
For example, when videos are collected with textual transcripts while framewise labels being unknown \cite{Richard-Viterbi:CVPR18,Ding:CVPR18,Chang:CVPR19,Li:ICCV19,Lu:ICCV21,Souri:PAMI21}, it is possible to annotate the labels of a small portion of videos to learn our model and apply it to obtain pseudo-labels for other videos then iteratively refine the model.

\begin{table}
   \centering
    \small
   \begin{tabular}{|cccc|ccc|c|c|}
   \hline
   $\mL_a$ & $\mL_c$ & $\mL_f$ & $\mL_{ts}$ & \multicolumn{3}{c|}{F1@\{10,25,50\}} & Edit & Acc \\ \hline
   $\checkmark$ & $\checkmark$ &  & $\checkmark$ & 73.4 & 68.2 & 57.3 & 73.3 & 70.9\\ \hline  
    & & $\checkmark$ & $\checkmark$ & 50.1 & 45.7 & 36.8 & 56.6 & 67.3 \\
   $\checkmark$ &  & $\checkmark$ & $\checkmark$ & 54.0 & 49.7 & 40.0 & 60.6 & 66.0 \\
   $\checkmark$ & $\checkmark$ & $\checkmark$ &  & 73.2& 69.3& 60.7& 72.7& 72.2  \\ \hline 
   $\checkmark$ & $\checkmark$ & $\checkmark$ & $\checkmark$ & \textbf{79.1} & \textbf{75.5}& \textbf{65.5} & \textbf{78.3} & \textbf{74.1} \\ \hline
   \end{tabular}
   \vspace{1mm}
   \caption{\small Effect of Different Losses.}
   \label{table:losses}
\end{table}

\subsection{Qualitative Results}
\label{sec:vis}
\begin{figure}
   \centering
   \includegraphics[width=0.45\textwidth]{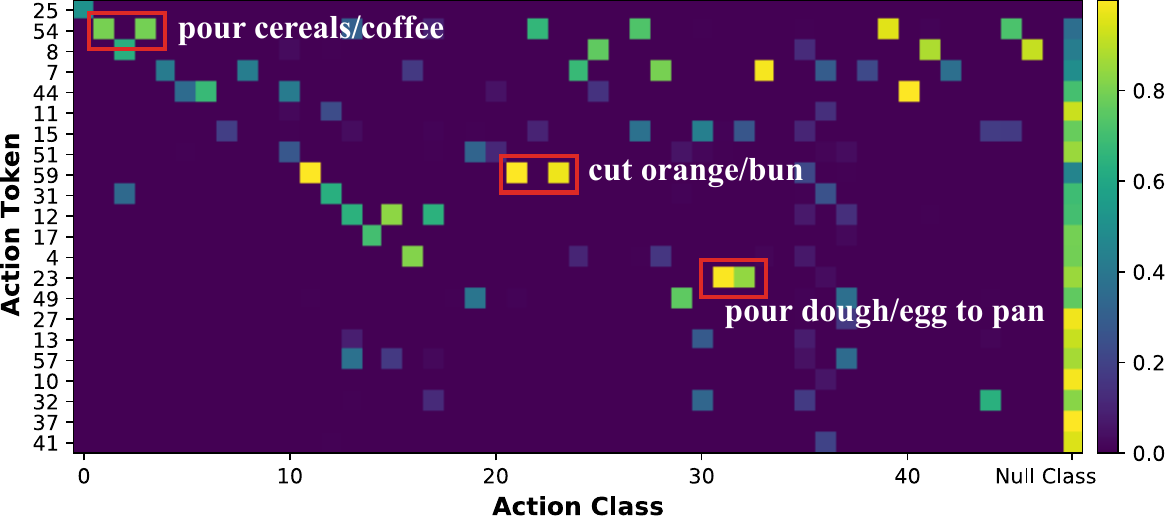}
   \captionof{figure}{\small Matching between Action Token and Action Class.}
   \label{fig:correlation}
\end{figure}

We study how our model assigns action tokens to encode the segments of different action classes. To do so, in Figure \ref{fig:correlation} for a subset of tokens on Breakfast, we show the frequency that the segments of one class is associated to a certain token (tokens are reordered to better highlight patterns).
Notice that, although we learned 60 tokens, more than the number of action classes, BIT does not simply assign one token per class but shares the token for similar classes. For example, token 54 is often related to \textit{pouring} actions; token 59 is related to \textit{cut} and token 23 is related to \textit{pouring things to pan}, showing \textbf{tokens have learned the semantic of the actions}.

\section{Conclusion}
\vspace{-3mm}
In this paper, we proposed BIT that i) efficiently learns a fixed-sized set of action tokens to explicitly represent action segments, ii) in parallel performs temporal modeling on action and frame levels, which allows us to outperform both frame-based and two-stage methods with a much lower computational cost. 
The design of the action tokens also enabled us to incorporate textual transcripts when they are available, thus can achieve higher segmentation accuracy with even less training data. 
We demonstrated the effectiveness of BIT with extensive experiments on four datasets along with various ablation studies. 
One limitation of BIT is that the number of action tokens is fixed for all videos, causing computation overhead for videos with only a few segments. In the future works, we will extend BIT to allow token pruning during inference.



{
\small
\bibliographystyle{ieeenat_fullname}
\bibliography{../biblio_bank/ehsan,
../biblio_bank/learning,
../biblio_bank/vision,
../biblio_bank/detection,
../biblio_bank/segmentation,
../biblio_bank/math,
../biblio_bank/procedurelearning,
../biblio_bank/action_understanding,
../biblio_bank/nlp}
}


\end{document}